\documentclass[lettersize,journal]{IEEEtran}

\usepackage{graphicx}
\graphicspath{{fig/}}

\usepackage{microtype}
\usepackage{url}
\usepackage{combelow}
\newcommand{\ii}[1]{{\footnotesize \textcolor{gray}{#1}}}

\usepackage{latexsym}
\usepackage[cmex10]{amsmath}
\usepackage{amssymb}
\usepackage{wasysym}
\usepackage{cancel}

\DeclareMathOperator*{\argmin}{arg\,min}

\newcommand{\defeq}{\triangleq}

\usepackage{booktabs}
\usepackage{multirow}
\usepackage{tabularx}
\newcommand{\mytable}{
	\centering
	\renewcommand{\arraystretch}{1.2}
}
\newcolumntype{C}{>{\centering\arraybackslash}X}
\newcolumntype{L}{>{\raggedright\arraybackslash}X}
\newcolumntype{R}{>{\raggedleft\arraybackslash}X}
\newcolumntype{P}[1]{>{\raggedright\arraybackslash}p{#1}}
\usepackage{siunitx}
\usepackage{etoolbox}                           
\newcommand{\ubold}{\fontseries{b}\selectfont}  
\robustify\ubold                                

\usepackage{cite}
\newcommand{\citet}[2]{#1~\cite{#2}}

\usepackage[prependcaption,textsize=scriptsize]{todonotes}
\definecolor{mycolor}{HTML}{FF6600}

\begin{document}

\title{Word Segmentation on Discovered Phone Units with Dynamic Programming and Self-Supervised Scoring}

\author{Herman Kamper
\thanks{E\&E Engineering, Stellenbosch University, South Africa.} 
}




\IEEEpubid{\copyright~2023 IEEE}

\maketitle

\begin{abstract}
    Recent work on unsupervised speech segmentation has used self-supervised models with phone and word segmentation modules      that are trained jointly.  This paper instead revisits an older approach to word segmentation: bottom-up phone-like unit discovery is performed first, and symbolic word segmentation is then performed on top of the discovered units (without influencing the lower level).  To do this, I propose a new unit discovery model, a new symbolic word segmentation model, and then chain the two models to segment speech.  Both models use dynamic programming to minimize segment costs from a self-supervised network with an additional duration penalty that encourages longer units.  Concretely, for acoustic unit discovery, duration-penalized dynamic programming~(DPDP) is used with a contrastive predictive coding model as the scoring network.  For word segmentation, DPDP is applied with an autoencoding recurrent neural as the scoring network.  The two models are chained in order to segment speech.      This approach gives comparable word segmentation results to state-of-the-art joint self-supervised segmentation models on an English benchmark. On French, Mandarin, German and Wolof data, it outperforms previous systems on the ZeroSpeech benchmarks. Analysis shows that the chained DPDP system segments shorter filler words well, but longer words might require some external top-down signal.
\end{abstract}


\section{Introduction}
\label{sec:introduction}

\IEEEPARstart{T}{he} {Zero-Resource Speech Team} at the 2012 JHU CLSLP workshop had a simple goal:
they wanted to develop an unsupervised approach that takes unlabelled speech and provides a full segmentation of the audio into word-like units~\cite{jansen+etal_icassp13}.
This could enable speech technology in very low-resource settings, and could be used in cognitive models that attempt to mimic how infants acquire language.
It seemed doable at the time because
there were several models that could perform full-coverage word segmentation from transcribed symbolic input (phoneme or phone sequences)~\cite{goldwater+etal_cognition09,johnson+goldwater_naacl09,elsner+etal_acl12}.
The team's idea was to use a clustering model to map speech to phone-like acoustic units---a Gaussian mixture model (GMM) was used---and to then apply symbolic segmentation on top of the discovered units.
Unfortunately, they found that this chained approach---where bottom-up acoustic unit discovery is performed independently of 
word segmentation---gave very poor results, with word token $F_1$ scores of around 5\%.
Compared to idealized symbolic input, the discovered units were 
too noisy and variable for a symbolic segmentation approach to pick up any meaningful signal~\cite{jansen+etal_icassp13}.

In the following decade,
several groups moved to joint speech segmentation models where bottom-up information from discovered acoustic units can influence top-down word segmentation and vice versa.
The hierarchical Bayesian HMM~\cite{lee+etal_tacl15} 
and the later
nonparametric Bayesian double-articulation analysers~\cite{taniguchi+etal_tcds16,nakashima+etal_frai19,okuda+etal_arxiv21}
are some examples.
Other groups tried to model higher-level units like syllables or words directly, circumventing the need for explicit phone modelling~\cite{rasanen+etal_interspeech15,kamper+etal_csl17,kamper+etal_asru17,wang+etal_icassp18}.
E.g., the Bayesian embedded segmental GMM (BES-GMM)~\cite{kamper+etal_csl17} performs probabilistic clustering and segmentation on fixed-dimensional representations of whole word-like segments, as does the related
embedded segmental $K$-means (ES-KMeans) model~\cite{kamper+etal_asru17}.
Much more recently, self-supervised neural networks using contrastive predictive coding (CPC)~\cite{vandenoord+etal_arxiv18} have been modified to include separate acoustic unit and word segmentation modules that are trained jointly~\cite{bhati+etal_interspeech21,cuervo+etal_arxiv21}.
These joint models give state-of-the-art word segmentation results (details in \S\ref{sec:related_work}).

Given that direct whole-word 
modelling~\cite{kamper+etal_csl17,kamper+etal_asru17} and joint phone and word modelling~\cite{bhati+etal_interspeech21,cuervo+etal_arxiv21} have proven to be such fruitful directions for speech segmentation, maybe the original premise of~\cite{jansen+etal_icassp13} was flawed?
This paper argues that their idea---where bottom-up acoustic unit discovery and word segmentation are performed separately---should be revisited.
Support for this 
comes from the big recent improvements in unsupervised unit discovery 
itself, specifically through 
self-supervised models that are coupled with a clustering step~\cite{dunbar+etal_interspeech19,dunbar+etal_interspeech20}.
These improved acoustic units (learned in a purely bottom-up fashion) could provide the signal necessary for symbolic word segmentation. 

\IEEEpubidadjcol

In this paper I specifically describe a duration-penalized dynamic programming (DPDP) procedure that combines a segment scoring function with a duration penalty.
Duration modelling is not a new idea in unsupervised word segmentation~\cite{mochihashi+etal_acl09,uchiumi+etal_acl15,kamper+etal_csl17,taniguchi+etal_tcds16,nakashima+etal_frai19,okuda+etal_arxiv21}, but coupling it with self-supervised neural scoring functions is. The proposed DPDP approach
can be used for either acoustic unit discovery or symbolic word segmentation, given an appropriate scoring network. 
I describe separate DPDP models for these two tasks, and then chain the two models in order to do speech segmentation.
For acoustic unit discovery, I apply DPDP with a CPC clustering model, 
an approach based on~\cite{roucos+dunham_milcom85,kamper+vanniekerk_interspeech21}.
For symbolic word segmentation, DPDP is applied with an autoencoding recurrent neural network (AE-RNN), an approach inspired by~\cite{elsner+shain_emnlp17}.
Chaining these two new DPDP models
gives similar performance to the joint self-supervised segmental~\cite{bhati+etal_interspeech21} and aligned~\cite{cuervo+etal_arxiv21} CPC models on English data.
The DPDP system
also achieves some of the best-reported word segmentation scores on the 
French, Mandarin, German and Wolof
ZeroSpeech 2017 and 2020 benchmarks~\cite{dunbar+etal_asru17,dunbar+etal_interspeech20},
where direct whole-word models have performed particularly well in the past~\cite{kamper+etal_asru17}.

This work can be seen as an extension of the conference paper~\cite{kamper+vanniekerk_interspeech21}, where some of the phone segmentation models in~\S\ref{sec:dpdp_cpc} were first proposed and evaluated on English.
Here I improve on these models by using better self-supervised scoring networks.
The main contributions of the paper are therefore as follows.
(1)~I propose a new model for acoustic unit discovery.
(2)~I propose a new approach for symbolic word segmentation.
(3)~These two DPDP-based models are then chained to segment speech. The chained system is applied to English and non-English data and compared to previous self-supervised and other state-of-the-art approaches.
(4)~I~analyse the combination of different acoustic unit discovery and symbolic segmentation models  (both DPDP and non-DPDP methods) for segmenting speech.
Taken together, the paper shows---for the first time---that 
using symbolic word segmentation on top of bottom-up discovered units can give comparable or better results compared to state-of-the-art joint self-supervised and direct whole-unit speech segmentation approaches.


The paper is structured as follows.
In \S\ref{sec:dpdp}, the general DPDP framework is introduced.
The new acoustic unit discovery and symbolic word segmentation approaches are then respectively presented in \S\ref{sec:dpdp_cpc} and \S\ref{sec:dpdp_aernn}.
These two sections have the same structure: I introduce the model, review related work, and give an intermediate evaluation on the model's specific task.
In \S\ref{sec:experiments} the two DPDP models are then chained, compared to previous speech segmentation approaches, and analysed.
As stated, I describe related work as it becomes relevant in the respective sections.
But it is worth starting by reviewing work that is relevant to the paper as a whole.

\section{Related Work}
\label{sec:related_work}


I briefly outline two specific approaches that I compare to throughout.
Both 
can be seen as extensions of contrastive predictive coding (CPC) and both can be used for either phone (\S\ref{sec:dpdp_cpc_intermediate}) or word segmentation from speech (\S\ref{sec:experiments_english}). 

Standard CPC speech models learn continuous features by trying to classify future observations (at different time-steps) from among a set of negative examples~\cite{vandenoord+etal_arxiv18}.
The idea is that a model would need to learn meaningful phonetic contrasts while being invariant to nuisance factors such as speaker.
Bhati et al.~\cite{bhati+etal_interspeech21} extend this by using a second segment-level CPC layer.
The segmental CPC (SCPC)
consists of a frame-level CPC module, a differentiable boundary detector operating on the learned features, and a segment-level CPC module operating on aggregated features from the lower layer.
The model is trained end-to-end using the combination of two losses: a next-frame classification loss for the lower-level CPC and a next-segment classification loss for the higher-level CPC.
For phone segmentation, peak detection is used to find points of high dissimilarity between the learned features.
For word segmentation, latent segment representations are compared.

Instead of treating classification independently for each future time-step as in standard CPC, the aligned CPC (ACPC) model of Chorowski et al.~\cite{chorowski+etal_interspeech21} outputs a sequence of predictions that are then aligned to future time-steps.
Since the model encourages piece-wise constant latent features, the idea is that changes in these features would correspond to phone boundaries.
To extend this model to word segmentation, the learned features are used for differentiable boundary detection and passed on to a higher-level ACPC module---very similar to the SCPC.
This multi-level ACPC (mACPC) is trained end-to-end on the combination of the lower- and higher-level ACPC losses~\cite{cuervo+etal_arxiv21}.
The mACPC gives some of the best unsupervised word segmentation results on the Buckeye benchmark (\S\ref{sec:experiments_english}).

Since both the SCPC and mACPC are trained end-to-end, phone discovery at the lower level can influence word discovery at the higher level and vice versa.
This is in contrast to the approach I propose, where phone discovery is performed without any influence from a word segmentation module---acoustic unit discovery is purely bottom-up.
My argument is not necessarily that the bottom-up approach is superior, but we will see that it is competitive with these 
CPC-based models.

\section{Duration-Penalized \\ Dynamic Programming~(DPDP)}
\label{sec:dpdp}

In this section I describe duration-penalized dynamic programming (DPDP) in its general form.
In the next sections I describe specific instances of DPDP-based models that can  be used for acoustic unit discovery or symbolic word segmentation.

\begin{figure}[!b]
    \centering
    \includegraphics[width=0.95\linewidth]{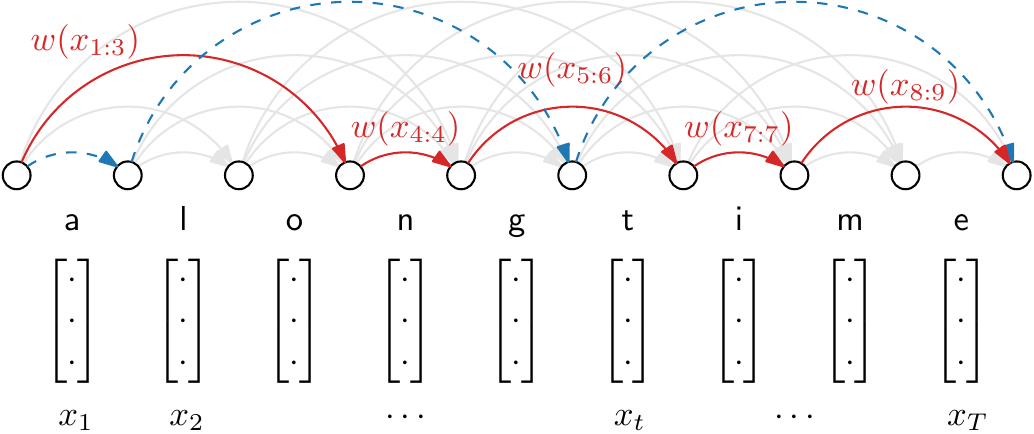}    
    \caption{The directed acyclic graph for DPDP with a maximum segment length of four.
        The segment cost $w(\cdot)$ should indicate how well that segment is modelled by a scoring network. E.g., on the dashed-blue path, $w(x_{2:5}) = w(\text{``long''})$ should be low if {``long''} is modelled well.
        The input can either be symbolic (shown as characters) or continuous (vectors), depending on the scoring network.
        Word segmentation is performed by finding the shortest path through the graph.}
    \label{fig:dpdp}
\end{figure}

We have an input sequence $X = x_{1:T} = \left(x_1, \ldots, x_T \right)$ that we want to segment. 
As illustrated in Figure~\ref{fig:dpdp}, this can either be a sequence of symbols, e.g.\ each $x_t \in \{ 1, \ldots, K \}$, or a sequence of speech features, e.g.\ $x_t \in \mathbb{R}^D$.
Now imagine we have some scoring network $w(\cdot)$ that takes a subsequence and gives a segment cost.
If $w(x_{a:b})$ is low, this indicates that the segment $x_{a:b}$ is modelled well by the scoring network.
Our goal is then to find the segmentation $S$ that gives the lowest overall cost when we sum up all the individual segment costs:
\begin{equation}
    \argmin_{S} \sum_{(a, b) \in S} w(x_{a:b})
    \label{eq:loss}
\end{equation}
We can solve this problem efficiently using dynamic programming.
This corresponds to finding the shortest path through a directed acyclic graph (DAG) with 
edge weights given by the scoring network. 
A segmentation $S$ can be specified as a sequence of $(\textrm{start}, \textrm{end})$ tuples.
The red-solid path in Figure~\ref{fig:dpdp} corresponds to the segmentation $S = \left( (1, 3), (4, 4) , (5, 6), (7, 7), (8, 9) \right)$, giving the result
``alo n gt i me''.
The blue-dashed path is a different segmentation with a different cost. 
The problem is 
to find the cheapest overall segmentation or, equivalently, the shortest path through the DAG.

To find this path using dynamic programming, we can
define
$$
    \alpha_t \defeq \min_{S_t} \sum_{(a, b) \in S_t} w(x_{a:b})
$$
as the cost for the optimal segmentation up to step $t$, where $S_t$ denotes a segmentation up to this intermediate point.
These forward variables can be calculated recursively:
\begin{equation}
    \alpha_t = \min_{j = 0}^{t - 1} \left\{ \alpha_j + w(x_{j + 1:t}) \right\}
    \label{eq:alphas}
\end{equation}
We start with $\alpha_0 = 0$ and calculate $\alpha_t$ for $t = 1, 2, \ldots, T$.
We keep track of the optimal choice ($\argmin$) for each $\alpha_t$ in~\eqref{eq:alphas}, and
the overall optimal segmentation is then obtained by starting
from the final step $t = T$ and moving backwards, repeatedly choosing the optimal boundary.

I have not explained the structure of the scoring network and will only do so in the next sections.
But it is worth briefly touching on one potential issue.
Depending on the properties of the scoring network, we could end up with a trivial solution.
One example is where the network always gives a very low cost when the input is a single symbol or single speech frame; this results in over-segmentation.
To deal with this, a penalty term is added to penalize shorter segments:
\begin{align}
    w(x_{a:b})
    &= w_{\textrm{seg}}(x_{a:b}) + \lambda \, w_{\textrm{dur}}(\textrm{dur}(x_{a:b})) \nonumber \\ 
    &= w_{\textrm{seg}}(x_{a:b}) + \lambda \, w_{\textrm{dur}}(b - a + 1)
    \label{eq:dp_weight}
\end{align}
Here I explicitly distinguish the segment cost $w_{\textrm{seg}}(\cdot)$ from the duration penalty $w_{\textrm{dur}}(\cdot)$ and 
 include a duration weight $\lambda$ that controls the relative importance of the two terms.

I refer to this procedure as duration-penalized dynamic programming (DPDP). Note that nothing in this description is new: it is really just a specific formulation of dynamic programming (see e.g.~\cite{mochihashi+etal_acl09,uchiumi+etal_acl15} for related duration-based models).
But the formulation here 
gives a common way to talk about and mathematically formulate the following two new models that are contributions of this work: one is used for acoustic unit discovery (\S\ref{sec:dpdp_cpc}) and the other for symbolic word segmentation~(\S\ref{sec:dpdp_aernn}). 
In the final experiments (\S\ref{sec:experiments}), I chain the two models to do 
word segmentation from speech.

\section{DPDP for Acoustic Unit Discovery}
\label{sec:dpdp_cpc}

Given a collection of unlabelled speech utterances, the goal of acoustic unit discovery is to learn a finite set of phone-like units representing the speech sounds that make up the language.
A model that can do this can also be used for unsupervised phone segmentation, predicting phone boundaries.
Several acoustic unit discovery and unsupervised phone segmentation models have been proposed~\cite{varadarajan+etal_acl08,gish+etal_interspeech09,lee+glass_acl12,michel+etal_arxiv16,wang+etal_interspeech17,ondel+etal_arxiv19}. Recently, large gains have been achieved by combining self-supervised neural networks with a clustering component.
Self-supervised speech models can produce continuous features that accurately capture phonetic contrasts while being invariant to nuisance factors (e.g.\ speaker)~\cite{vandenoord+etal_arxiv18,chung+etal_interspeech19,vanniekerk+etal_interspeech20,vanniekerk+etal_interspeech21}.
To obtain discrete units, a self-supervised model can be coupled with a vector quantization~(VQ) module, either as part of the model itself or by introducing a clustering step after training~\cite{chorowski+etal_taslp19,vanniekerk+etal_interspeech20,vanniekerk+etal_interspeech21,baevski+etal_iclr20,nguyen+etal_sas20,hsu+etal_taslp21}.
Despite improvements in phone discrimination tasks~\cite{dunbar+etal_interspeech20}, these approaches still encode speech at a much higher bitrate than true phone sequences~\cite{coupe+etal_sciadv19}.
The main reason is that the assignment of features to VQ codebook vectors is done independently for each speech frame at a fixed rate.
In reality, adjacent frames are likely to belong to the same speech unit, and these units occur at a variable rate.
The SCPC and ACPC (\S\ref{sec:related_work}) try to address this problem through intermediate boundary detection.
Below I propose a different approach.

\subsection{DPDP on vector-quantized self-supervised speech features}

The approach here constrains a VQ model so that contiguous feature vectors are assigned to the same code, resulting in a variable-rate encoding of the input speech.
I first explain the approach by itself before formulating it as an instance of DPDP. 

\begin{figure}[!b]
    \centering
    \includegraphics[width=0.9\linewidth]{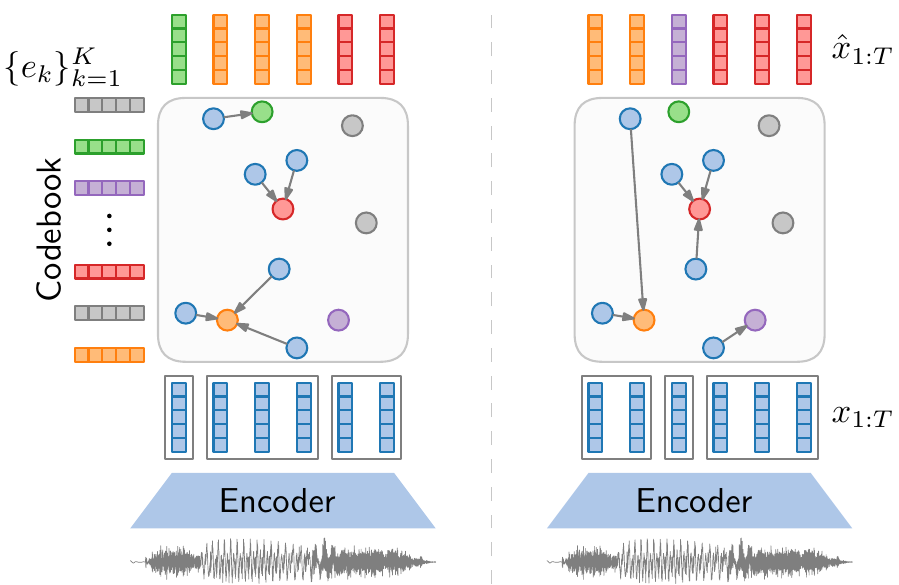}
    \caption{Two segmentations of an utterance.
        The features in each segment are assigned to the same code.
        The left will have a smaller sum of squared distances between the features and their assigned codes than the right.}
    \label{fig:dpdp_cpc}
\end{figure}

Given input speech, continuous feature vectors are first extracted 
using an encoder from a self-supervised speech model.
Let us denote these features as $x_{1:T}$, where each $x_t \in \mathbb{R}^D$ is a learned $D$-dimensional feature vector.
We then solve a constrained optimization problem through which this sequence is divided into segments.
Figure~\ref{fig:dpdp_cpc} illustrates this, showing two possible segmentations.
All the features within a segment are assigned to the same code from a VQ codebook $\{ e_k \}_{k = 1}^K$, with each $e_k \in \mathbb{R}^D$.
The overall cost for a particular segmentation is the sum of the squared distances of the features and the representative code of each segment.
In Figure~\ref{fig:dpdp_cpc}, this cost corresponds to adding up the squared lengths of the arrowed lines.
Our objective is to find the segmentation $S$ that minimizes the overall summed distances:
\begin{equation}
\argmin_{S} \sum_{(a, b) \in S} \  \sum_{x_t \in x_{a:b}} || x_t - \hat{x}_t ||^2
\label{eq:seg_squared_cost}
\end{equation}
where $\hat{x}_t$ is the codebook vector to which the segment $x_{a:b}$ is assigned, and the segmentation $S$ is specified as in \S\ref{sec:dpdp}.
One problem with this overall cost is that the best segmentation will always place each $x_t$ in its own segment, assigning it to the code closest to it.
Additional constraints are therefore required: I specifically introduce a duration penalty to encourage longer but fewer segments. To incorporate this penalty, let us now frame this 
as an instance of DPDP (\S\ref{sec:dpdp}).

To get to the overall cost in~\eqref{eq:seg_squared_cost} (without the duration penalty), we need the following segment cost:
$$
w_{\textrm{seg}} (x_{a:b}) = \min_{k = 1}^K \sum_{x_t \in x_{a:b}} || x_t - e_k ||^2
$$
We then add a duration penalty $w_{\textrm{dur}}(l)$.
There are several options. One is to use a probabilistic prior over acoustic unit duration (more on this in \S\ref{sec:dpdp_aernn_model}).
Here I use a simpler linear penalty, based on~\cite{chorowski+etal_neuripspgr9}: $w_{\textrm{dur}}(l) = -l + 1$.
Following the DPDP formulation, the combined weight is then as in~\eqref{eq:dp_weight}, and the best segmentation in~\eqref{eq:loss} is obtained by recursively calculating~\eqref{eq:alphas}.

\subsection{Related work}

The above approach was first introduced in~\cite{kamper+vanniekerk_interspeech21}, where it was specifically coupled with VQ-VAE and VQ-CPC scoring networks with a VQ layer that is learned during self-supervised training.
In the intermediate evaluation below, I show that better performance can be achieved by applying DPDP on a large CPC model that uses $K$-means clustering to obtain VQ representations 
after self-supervised training. 

A significant shortcoming of~\cite{kamper+vanniekerk_interspeech21} was that it failed to connect the approach to much older work done at BBN in the 1980s.
Building on an earlier heuristic approach~\cite{roucos+etal_icassp82}, Roucos and Dunham~\cite{roucos+dunham_milcom85} proposed an approach for low-bitrate speech coding: linear prediction speech frames are jointly segmented and quantized using a codebook through a dynamic programming procedure.
Because they used an alignment-based segment cost, they did not have to include an explicit duration penalty as I do here.
The only other difference is that their approach operates directly on speech frames, while the one here operates on learned self-supervised features.

The DPDP approach in this section is also a generalization of the more recent work of Chorowski et al.~\cite{chorowski+etal_neuripspgr9}.
Instead of using a duration penalty, they enforce a prespecified number of segments into which a sequence needs to be divided.
It can actually be shown that, if we set the duration penalty $w_{\textrm{dur}}(l) = -l + 1$ as above and have a unique $\lambda$ for each utterance, their formulation is the dual of the one presented here.
This only holds for this particular duration penalty and not in general.
Moreover, their implementation 
used a greedy approximation; it was shown in~\cite{kamper+vanniekerk_interspeech21} that better phone segmentation results are achieved when doing full dynamic programming.

\subsection{Intermediate evaluation: Unsupervised phone segmentation}
\label{sec:dpdp_cpc_intermediate}

I present a brief intermediate evaluation here.
The goal is not to achieve state-of-the-art performance, but rather to verify that the DPDP method can give reasonable phone segmentation results (before using it as input for symbolic word segmentation).
Complete experimental details are given in \S\ref{sec:experimental_setup}, but in short, here I follow the experimental setup of~\cite{kreuk+etal_interspeech20} on the English Buckeye speech corpus.
In addition to phone boundary precision, recall and $F_1$, I also report over-segmentation (OS): how many fewer/more boundaries are proposed compared to the ground truth. 
The ideal is 0\%. 
$F_1$ is not always sensitive enough to the trade-off between recall and over-segmentation, which motivates the $R$-value~\cite{rasanen+etal_interspeech09}: it gives a perfect score (100\%) when a method has perfect recall
and perfect~OS.

Table~\ref{tbl:phoneseg_test} gives the results for five state-of-the-art approaches~\cite{michel+etal_arxiv16,wang+etal_interspeech17,kreuk+etal_interspeech20,bhati+etal_interspeech21,cuervo+etal_arxiv21}.
DPDP is used with three self-supervised models. 
The VQ-VAE and VQ-CPC scoring networks are from~\cite{vanniekerk+etal_interspeech20}.
Both networks' encoders take log-mel spectrograms as input, downsample it by two, and discretize the feature vectors using a VQ layer with 512 codes.
The networks (including the VQ layers) are trained on the 15-hour English training set from ZeroSpeech 2019~\cite{dunbar+etal_interspeech19}.
The third DPDP system uses the CPC-big model from~\cite{nguyen+etal_sas20}, trained on LibriLight unlab-6k~\cite{kahn_icassp20}.
CPC-big takes raw speech input.
After training the network, $K$-means is applied to features extracted from the second layer in its four-layer LSTM context network; features are extracted from the LibriSpeech train-clean-100 set. 
The result is a VQ codebook with 50 codes.
Since the structure of the codebooks differ in the three DPDP systems, the duration weight in~\eqref{eq:dp_weight} is set individually: $\lambda = 3$ for the DPDP VQ-VAE, $\lambda = 400$ for DPDP VQ-CPC, and $\lambda = 2$ for DPDP CPC+$K$-means.
These weights were 
tuned on Buckeye development data (see \S\ref{sec:experimental_setup}).

\begin{table}[!t]
    \mytable
    \caption{Intermediate phone boundary segmentation results (\%) on Buckeye test data for state-of-the-art models and DPDP segmentation.}
    \begin{tabularx}{\linewidth}{@{}lCCCS[table-format=2.1]S[table-format=2.1]@{}}
        \toprule
        Model & Prec. & Rec. & $F_1$ & {OS} & {$R$-val.} \\
        \midrule
        \textit{\underline{Unsupervised:}} \\
        \addlinespace
        GRU next-frame prediction~\cite{michel+etal_arxiv16} & 69.3 & 65.1 & 67.2 & -6.1 & 72.1 \\
        GRU gate activation~\cite{wang+etal_interspeech17} & 69.6 & 72.6 & 71.0 & -4.1 & 74.8 \\
        Self-sup.\ contrastive~\cite{kreuk+etal_interspeech20} & 75.8 & 76.9 & 76.3 & \ubold -1.4 &   79.7 \\
        SCPC~\cite{bhati+etal_interspeech21} & \ubold 76.5 & 78.7 & \ubold 77.6 & {-} & \ubold 80.7 \\
        ACPC~\cite{cuervo+etal_arxiv21} & 74.4 & 76.3 & 75.3 & {-} & 78.7 \\
        \addlinespace
        Merged CPC+$K$-means (no DP) & 36.9 & \ubold 97.2 & 53.5 & 164.5 & -40.5 \\
        \addlinespace
        DPDP VQ-CPC & 69.6 & 73.4 & 71.5 & 5.4 & 75.1 \\
        DPDP VQ-VAE & 70.8 & 85.6 & 77.5 & 20.9 & 74.8 \\
        DPDP CPC+$K$-means & 73.2 & 77.7 & 75.4 & 6.2 & 78.3 \\
        \addlinespace
        \textit{\underline{Supervised:}} \\
        \addlinespace
        LSTM~\cite{franke+etal_itg16} & 87.8 & 83.3 & 85.5 & 5.4 & 87.2 \\
        LSTM structured loss~\cite{kreuk+etal_icassp20} & 85.4 & 89.1 & 87.2 & -4.1 & 88.8 \\		
        \bottomrule
    \end{tabularx}
    \label{tbl:phoneseg_test}
\end{table}

Table~\ref{tbl:phoneseg_test} shows that the DPDP approach outperforms~\cite{michel+etal_arxiv16,wang+etal_interspeech17}, but performs slightly worse on most metrics compared to the recent state-of-the-art unsupervised phone segmentation approaches~\cite{kreuk+etal_interspeech20,bhati+etal_interspeech21,cuervo+etal_arxiv21}.
In terms of precision, $F_1$, OS and $R$-value, the DPDP systems also all outperform a system where no duration penalty is applied (no DP), i.e.\ repeated cluster indices are simply merged. This no-DP approach results in severe over-segmentation.
Of the three DPDP systems, the CPC+$K$-means model (here used as a DPDP scoring network for the first time) achieves the best precision and $R$-value, while the DPDP VQ-VAE gives better recall and $F_1$ scores.
Because of these intermediate results, I do not report results with the VQ-CPC in \S\ref{sec:experiments}.

\section{DPDP for Unsupervised Word Segmentation from Symbolic Input}
\label{sec:dpdp_aernn}

The goal in unsupervised word segmentation from symbolic input is to break up an input sequence (normally of phonemes or phones) into subsequences representing words.
My aim here is specifically to develop a model that can operate on the noisy symbolic sequences from an acoustic unit discovery model (such as the ones in \S\ref{sec:dpdp_cpc}).
The new model that I introduce is
a simplified version of~\cite{elsner+shain_emnlp17}, as explained below in~\S\ref{sec:dpdp_aernn_related_work}.

\subsection{DPDP autoencoder recurrent neural network (AE-RNN)}
\label{sec:dpdp_aernn_model}

This approach is again an instance of DPDP.
Let us say we have an autoencoding recurrent neural network (AE-RNN): an encoder-decoder model that takes a sequence of symbols as input, summarizes these into a single embedding vector using an encoder RNN, conditions a decoder RNN on the embedding, and finally tries to produce an output sequence that matches the input.
Such a model could be seen as compressing the input sequence into a fixed-dimensional embedding, with the decoder then acting as a decompression algorithm.
After training, the AE-RNN could give an accurate reconstruction of some inputs, while for some other inputs the reconstruction could be bad.

\begin{figure}[!b]
    \centering
    \includegraphics[width=0.99\linewidth]{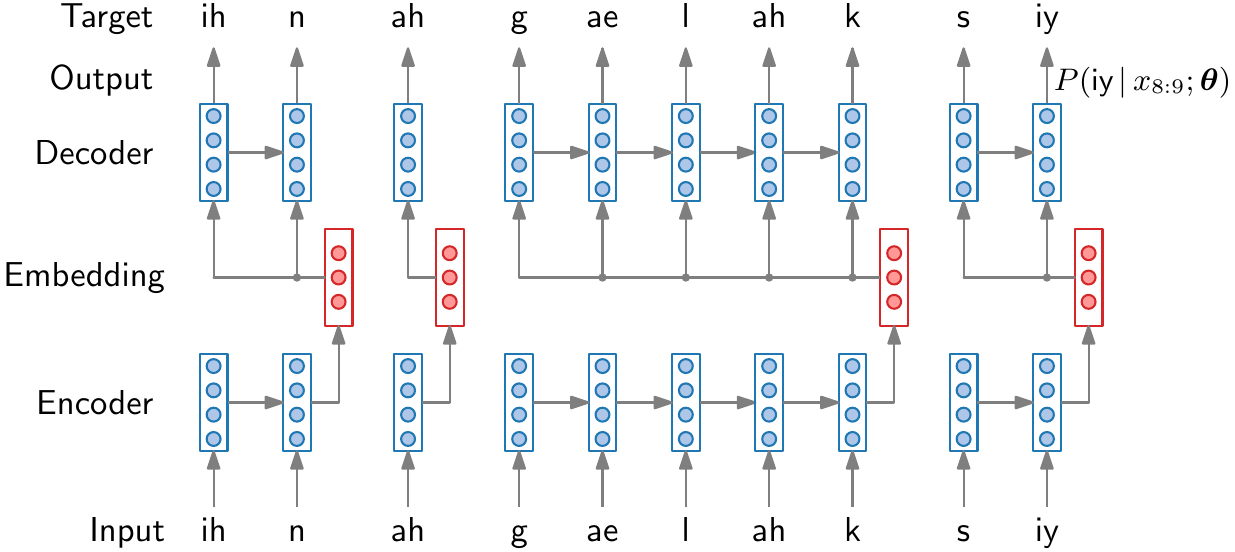}
    \caption{An example where a symbolic sequence is divided into four segments, each 
    processed through an autoencoding RNN (AE-RNN). The DPDP AE-RNN considers different segmentations and selects the one with the lowest combined negative log likelihood and duration cost.}
    \label{fig:dpdp_aernn}
\end{figure}

We can use this property to perform segmentation of an input symbol sequence $x_{1:T}$. Our goal is to divide the sequence into segments, each of which is accurately modelled by the AE-RNN.
Figure~\ref{fig:dpdp_aernn} illustrates this for one possible segmentation.
Here each $x_t \in \{k\}_{k = 1}^K$ takes on a discrete value.
Each of the four segments in Figure~\ref{fig:dpdp_aernn} can be scored based on its negative log likelihood according to the AE-RNN:
$$
w_{\textrm{seg}}(x_{a:b}) = - \sum_{x_t \in x_{a:b}} \log P(x_t | x_{a:b}; \boldsymbol{\theta})
$$
where $P(x_t | x_{a:b}; \boldsymbol{\theta})$ is the $t^{\text{th}}$ output of the decoder (obtained after a softmax layer) when the encoder is presented with segment $x_{a:b}$. $\boldsymbol{\theta}$ is the parameters of the AE-RNN.
An overall segmentation cost is obtained by summing the individual segment costs.
A different segmentation in Figure~\ref{fig:dpdp_aernn} will lead to a different overall cost, and we want to find the minimum.

There is a trivial solution: an AE-RNN that places a high output probability only on the first input symbol would give a result where every symbol is in its own segment.
I therefore again introduce a duration penalty $w_{\textrm{dur}}(l)$ and formulate this approach as an instance of DPDP  (\S\ref{sec:dpdp}).

There are different options for $w_{\textrm{dur}}(l)$.
In the main experiments in \S\ref{sec:experiments}, I use a 
linear penalty for the DPDP AE-RNN. 
But
for the intermediate evaluation here (\S\ref{sec:dpdp_aernn_eval}), I use a fixed probability mass function from a truncated gamma density (truncated at 50 symbols).
This DPDP can correspond to a valid probabilistic model, but then we also need to model the number of segments in an utterance:
I use a geometric distribution and set $\lambda$ = 1 in~\eqref{eq:dp_weight}. 
With these settings, this DPDP corresponds to a hidden semi-Markov model (HSMM)~\cite{murphy_notes02} with a fixed duration distribution for all states and the same emission 
distribution (parametrized by 
the AE-RNN).\footnote{HSMMs~\cite{murphy_notes02} are probabilistic models where the duration that you stay in a hidden state is explicitly modelled.
If we use a duration penalty that corresponds to a valid probability distribution, set $\lambda = 1$, model transitions between states (e.g.\ uniform), and model the number of segments in an utterance, then the DPDP AE-RNN \textit{is} an HSMM with a fixed duration distribution and emission distributions parametrized by the AE-RNN (a parametrization not seen before, as far as I know).
This is the variant of the DPDP AE-RNN used in the intermediate evaluation here (\S\ref{sec:dpdp_aernn_eval}).
However, the more general formulation of the DPDP AE-RNN given in this section is not bound by the constraint of a valid probabilistic duration prior, and indeed the model used in the speech segmentation experiments (\S\ref{sec:experiments}) uses a simple linear penalty (this gave better performance on development data).
Also note that the ES-KMeans~\cite{kamper+etal_asru17} and BES-GMM~\cite{kamper+etal_csl17} that I compare to in \S\ref{sec:experiments} can also be formulated as HSMMs.
HSMMs are therefore widespread, but the details of how they are parametrized are crucial and can lead to very different results.}

Up to now I implicitly assumed that we have a trained AE-RNN, but in reality we will need to fit its parameters $\boldsymbol{\theta}$.
One way would be to start with a random segmentation, train the AE-RNN on these random segments, find the optimal segmentation under this AE-RNN, update the AE-RNN, re-segment, and so on.
Another approach is that of~\cite{elsner+shain_emnlp17}: instead of only considering the single best segmentation, 
they probabilistically
sample likely segmentations. This could help better explore the segmentation space.
Instead of either of these approaches, I found that a simple approach gave robust results: simply train the AE-RNN on the complete full-length utterances in the dataset.
This AE-RNN is trained once and then fixed as the DPDP scoring network.
This worked better than iterative refinement in developmental experiments.

\subsection{Related work}
\label{sec:dpdp_aernn_related_work}

Word segmentation of symbolic sequences has a long history, particularly in the cognitive science community where these models are used to investigate how infants learn to segment words and discover the lexicon of their native language~\cite{goldwater+etal_cognition09}.
Most models are applied on transcribed phonemic input where every word is represented by the same sequence of phonemes.\footnote{Some studies also consider variation in word pronunciation, taking phonetic input~\cite{elsner+etal_emnlp13}.
Although not evaluated in the intermediate evaluation in this section, the DPDP AE-RNN can also deal with such variation, as implicitly illustrated in the main experiments in \S\ref{sec:experiments} where I apply it on noisy discovered units.}

Some of the first approaches relied on the assumption that the transitions between symbols within a word are more predictable than across words~\cite{brent_ml99}.
Based on this, a word boundary can be predicted when the transition probability between two symbols dips below a threshold~\cite{saksida+etal_develsci17}.
Another approach is to explicitly model word units.
The hierarchical Dirichlet process model~\cite{goldwater+etal_cognition09} uses a bigram language model over inferred word tokens with priors to encourage predictable word sequences and a small vocabulary.
A related nonparametric 
Bayesian approach is the adaptor grammar~\cite{johnson+goldwater_naacl09}, which assumes that a corpus is generated from a set of re-write rules that can be learned probabilistically; below I specifically use the ``colloc'' variant, which includes bigram-like rules for co-occurring word units. 

More recently, researchers have turned to neural sequence models, e.g.~\cite{kawakami+etal_acl19}.
The RNN memory segmenter of Elsner and Shain~\cite{elsner+shain_emnlp17} is particularly relevant here, since the DPDP AE-RNN can be seen as a simplification of this model.
As explained above, I train the AE-RNN once on the full-length utterances in the dataset and then find the single best segmentation using~\eqref{eq:loss} and~\eqref{eq:alphas}.
Instead of taking the single best segmentation,~\cite{elsner+shain_emnlp17} iteratively refines the AE-RNN by probabilistically sampling boundaries.
To help with the search problem, they train a separate RNN to propose boundaries, and use its smoothed output to sample boundaries.
Below I show that their approach performs marginally better on phonemic data; but the DPDP AE-RNN was easier to tune in the chained approach  where it is coupled with an acoustic unit discovery model (\S\ref{sec:experiments}).

\subsection{Intermediate evaluation: Unsupervised word segmentation of phoneme sequences}
\label{sec:dpdp_aernn_eval}

In this intermediate evaluation I verify that the DPDP AE-RNN gives reasonable word segmentation results when applied to phonemic sequences.\footnote{I release a separate repository for the DPDP AE-RNN to reproduce the experiments in this section: {\tt \scriptsize \url{http://github.com/kamperh/dpdp_aernn}}.}
Again the goal is not state-of-the-art results (although I compare to established systems).
I specifically consider results on the Brent corpus~\cite{brent_ml99}, a standard benchmark for symbolic word segmentation.

I tune the hyperparameters of the DPDP AE-RNN on the first 1000 utterances of the corpus.
This includes the parameters for the gamma-based duration and geometric sentence length priors (\S\ref{sec:dpdp_aernn_model}).
The AE-RNN consists of a three-layer GRU encoder and a single-layer GRU decoder, all with 200-dimensional hidden vectors; the latent embedding has 25 dimensions.
The model is trained and evaluated on the full 9790-utterance corpus, as is the practice in other work~\cite{elsner+shain_emnlp17}; results are very similar when removing the first 1000 utterances.

\begin{table}[!t]
    \mytable
    \caption{Intermediate word segmentation results (\%) on the Brent corpus (transcribed phoneme sequences) for previous symbolic word segmentation methods and the DPDP AE-RNN.}
    \begin{tabularx}{\linewidth}{@{}lCS[table-format=2.0]CS[table-format=2.0]@{}}
        \toprule
        & \multicolumn{3}{c}{{Word boundary}} & {Token} \\
        \cmidrule(lr){2-4} \cmidrule(l){5-5}
        Model & Prec. & {Rec.} & $F_1$ & {$F_1$} \\
        \midrule
        Every phoneme as a word (no DP) & 27 & 100 & 43 & 3 \\
        \addlinespace
        Transition probability~\cite{saksida+etal_develsci17} & 59 & 71 & 64 & 47 \\
        Hierarchical Dirichlet process~\cite{goldwater+etal_cognition09} & \ubold 90 & 74 & \ubold 87 & 74 \\
        Adaptor grammar~\cite{johnson+goldwater_naacl09} & - & {-} & - & \ubold 88 \\
        RNN memory segmenter~\cite{elsner+shain_emnlp17} & 81 & \ubold 85 & 83 & 72 \\
        \addlinespace
        DPDP AE-RNN & 78 & \ubold 85 & 81 & 69 \\
        \bottomrule
    \end{tabularx}
    \label{tbl:symbolic_wordseg}
\end{table}

Table~\ref{tbl:symbolic_wordseg} shows word segmentation results.
Apart from word boundary scores, where each boundary decision is evaluated separately, the table also shows the word token $F_1$ score, which requires both boundaries of a word to be correctly predicted without any intermediate boundary proposals.
This metric therefore also implicitly penalizes over-segmentation.
We see that the DPDP AE-RNN gives similar or slightly worse performance compared to~\cite{elsner+shain_emnlp17}, on which it is based.
Although the adaptor grammar~\cite{johnson+goldwater_naacl09} gives the best token $F_1$ score here, we will see in \S\ref{sec:experiments} that this method performs much worse when applied on discovered acoustic units---so better performance  on phonemic sequences does not necessarily translate to better downstream speech segmentation results.

\section{Experimental Setup}
\label{sec:experimental_setup}

Our main experiments involve the task of unsupervised word segmentation from speech.
Since this
is an unsupervised task, a model is typically trained and evaluated on the same data~\cite{versteegh+etal_interspeech15}.
This means that developmental experiments need to be performed carefully, and 
 I follow the practice of~\cite{kamper+etal_taslp16,dunbar+etal_asru17}:
Hyperparameters are chosen by training and evaluating on a development dataset.
All hyperparameters are then fixed.
For testing, the model is then trained and evaluated on a different set (potentially from another language) without any changes to any hyperparameters from the developmental experiments.

Development is performed on a 3.5-hour development set from Buckeye~\cite{pitt+etal_speechcom05}.
For evaluation on English,
testing is then done on the 5-hour test set from Buckeye.
There is no speaker overlap between these two sets.
I then also apply the DPDP system to four 
completely unseen languages using the
French, Mandarin, German and Wolof
datasets from the ZeroSpeech 2017 challenge~\cite{dunbar+etal_asru17}, respectively containing
24, 2.5, 25 and 10
hours of active speech.
This data was also used in ZeroSpeech 2020.
I report word boundary precision, recall, $F_1$, OS and $R$-value, as well as word token $F_1$, 
all with a 20~ms tolerance. 

The goal is to see whether DPDP-based symbolic word segmentation (\S\ref{sec:dpdp_aernn}) can be applied on top of DPDP-discovered acoustic units (\S\ref{sec:dpdp_cpc}) in order to segment speech.
For acoustic unit discovery, I use the DPDP models already described in \S\ref{sec:dpdp_cpc_intermediate}.
All the scoring networks (e.g.\ 
VQ-VAE and CPC+$K$-means) are pretrained on English data and the hyperparameters are set exactly as explained in that subsection.

For word segmentation, I use the DPDP AE-RNN.
Here it is set up slightly differently from the one in the intermediate evaluation in \S\ref{sec:dpdp_aernn_eval}: the one here has a 10-dimensional symbol embedding layer, a single 500-dimensional GRU encoder layer, a 50-dimensional latent embedding, and a single 500-dimensional GRU decoder.
Instead of using a probabilistic duration penalty and sequence length prior, I use the simple duration penalty $w_{\textrm{dur}}(l) = -l + 1$ with a weight of $\lambda = 3$.
This is also the duration penalty used in the DPDP unit discovery models.
This simpler DPDP AE-RNN gave better development performance when applied on discovered units.
The AE-RNN is trained with Adam optimization~\cite{kingma+ba_iclr15} using a learning rate of $1\cdot10^{-3}$ for 1500 steps on the full tokenized utterances.
The combined time for training and doing forward DPDP inference in the AE-RNN on the Buckeye development data is roughly 15 minutes on a machine with an NVIDIA GeForce RTX~3070 GPU and a single 2.5~GHz CPU. This is apart from the DPDP forward inference time for acoustic unit discovery, which is roughly 12 minutes on its own.
I release code at {\tt \small \url{http://github.com/kamperh/vqwordseg}}.

\section{Experiments: DPDP for Unsupervised Word Segmentation from Speech}
\label{sec:experiments}

The goal is to see how bottom-up  acoustic unit discovery followed by symbolic word segmentation 
compare to other approaches.
My specific new proposal is to chain the DPDP models from \S\ref{sec:dpdp_cpc} and \S\ref{sec:dpdp_aernn}.
I first compare this chained DPDP system to state-of-the-art systems on English~(\S\ref{sec:experiments_english}).
Then I consider different combinations of symbolic segmentation with acoustic unit discovery models~(\S\ref{sec:experiments_chained}). Finally I compare to other existing systems on non-English data~(\S\ref{sec:experiments_nonenglish}).

\subsection{Comparing to joint self-supervised and direct whole-unit approaches on English}
\label{sec:experiments_english}

\begin{table}[!b]
	\mytable
	\caption{Unsupervised word segmentation results (\%) on Buckeye test data.}
	\begin{tabularx}{\linewidth}{@{}L@{\ }S[table-format=2.1]S[table-format=2.1]S[table-format=2.1]S[table-format=2.1]S[table-format=2.1]S[table-format=2.1]@{}}
		\toprule
		& \multicolumn{5}{c}{{Word boundary}} & {Token} \\
		\cmidrule(r){2-6} \cmidrule(l){7-7}
        {Model} & {Prec.}  & {Rec.}  & {$F_1$} & {OS}  & {$R$-val.} & {$F_1$} \\
		\midrule
		\ii{1:} Adaptor gr.\ 
        on GMM~\cite{jansen+etal_icassp13} & 15.9 & 57.7 & 25.0 & 261.5 & -139.9 & 4.4 \\
        \addlinespace
		\ii{2:} $K$-means on syll.~\cite{rasanen+etal_interspeech15} & 27.7 & 28.9 & 28.3 & \ubold 4.5 & 37.7 & 19.3 \\
		\ii{3:} ES-KMeans~\cite{kamper+etal_asru17} & 30.3 & 16.6 & 21.4 & -45.1 & 39.1 & 19.2 \\
		\ii{4:} BES-GMM~\cite{kamper+etal_csl17} & 31.5 & 12.4 & 17.8 & -60.5 & 37.2 & 18.6 \\
		\addlinespace
		\ii{5:} SCPC~\cite{bhati+etal_interspeech21} & 34.8 & 31.0 & 32.8 & -10.8 & 44.5 & {-} \\
		\ii{6:} mACPC~\cite{cuervo+etal_arxiv21} & \ubold 42.1 & 30.3 & 35.1 & -26.2 & \ubold 47.4 & {-} \\
		\addlinespace
        \ii{7:} Merged CPC+$K$-means & 9.1 & \ubold 94.5 & 16.6 & 936.6 & -701.4 & 1.5 \\
        \addlinespace
		\ii{8:} DPDP AE-RNN on \mbox{\qquad DPDP CPC+$K$-means} & 35.3 & 37.7 & \ubold 36.4 & 6.7 & 44.3 & \ubold 25.0 \\
		\bottomrule
	\end{tabularx}
	\label{tbl:wordseg_buckeye}
\end{table}

Table~\ref{tbl:wordseg_buckeye} shows word segmentation results on the English Buckeye test data, comparing the chained DPDP system (row~8) to existing approaches.

Like the DPDP system, row~1 is a system where bottom-up acoustic unit discovery is followed with symbolic word segmentation: a 50-component GMM is trained on unlabelled speech, the speech is encoded according to the most probable component for each MFCC frame, repeated components are merged, and the resulting tokenization is segmented with an adaptor grammar~\cite{johnson+goldwater_naacl09}.
This is representative of the approaches followed in~\cite{jansen+etal_icassp13} (see \S\ref{sec:introduction}).
The DPDP system (row~8) outperforms this method on all metrics.
This therefore represents a large improvement in bottom-up discovery followed by segmentation---the methodology shared by the two systems.

Instead of explicitly learning phone-like acoustic units, the systems in rows~2 to 4 all try to model higher-level units directly without an explicit lower-level acoustic unit layer.
They either treat syllables~\cite{rasanen+etal_interspeech15} or words~\cite{kamper+etal_asru17,kamper+etal_csl17} as the basic modelling unit (see \S\ref{sec:introduction}).
While these approaches outperform the adaptor grammar on GMM units (row~1), the DPDP system (row~8) outperforms all these direct whole-unit models across all 
metrics (except for~\cite{rasanen+etal_interspeech15} in row~2 giving a better OS score). 

Finally I compare to two recent state-of-the-art CPC-based models (rows~5 and~6), described in \S\ref{sec:related_work}.
In contrast to the row 2 to 4 systems, these models have separate modules for acoustic unit learning and word segmentation which are learned jointly. This allows top-down information from word segmentation to influence bottom-up acoustic unit learning~\cite{bhati+etal_interspeech21,cuervo+etal_arxiv21}.
The DPDP system performs better on word boundary recall, 
$F_1$ and OS, but the CPC-based systems achieve better precision and $R$-values (word token $F_1$ was not reported in these papers).

As a sanity check, I also give the result that would be obtained if no duration penalty was used on top of the self-supervised CPC+$K$-means model (row 7), i.e.\ repeated cluster indices are simply merged and then treated as words.
As was the case in the intermediate phone segmentation experiments (Table~\ref{tbl:phoneseg_test}), this no-DP approach over-segments heavily, resulting in high boundary recall but poor precision.

In summary, the DPDP system outperforms bottom-up and direct whole-unit modelling, and performs similarly to joint self-supervised CPC-based models on an English benchmark.

\subsection{Symbolic word segmentation on discovered acoustic units}
\label{sec:experiments_chained}

\begin{table}[!b]
    \mytable
    \caption{Word token $F_1$ scores (\%) on Buckeye development data using different combinations of acoustic unit segmentation methods (rows) and different word segmentation approaches (columns).}
    \begin{tabularx}{\linewidth}{@{}l@{\quad}l@{}CCC@{}}
        \toprule
        & & \multicolumn{3}{c}{Symbolic word segmentation}  \\
        \cmidrule{3-5}
        & & Trans.\ prob.~\cite{saksida+etal_develsci17} & Adaptor gram.~\cite{johnson+goldwater_naacl09} & DPDP \mbox{AE-RNN} \\
        \midrule
        \multirow{4}{*}{\parbox{0.9cm}{Acoustic unit \ \  discovery}}
        & MFCC+GMM & 5.8 & 4.7 & 17.2 \\
        & Merged CPC+$K$-means (no DP) & 6.5 & 5.5 & 22.5 \\
        & DPDP VQ-VAE & 9.4 & 5.3 & 16.4 \\
        & DPDP CPC+$K$-means & 9.7 & 8.7 & \ubold 24.1 \\
        \bottomrule
    \end{tabularx}
    \label{tbl:wordseg_val}
\end{table}	

It is clear that
symbolic segmentation on top of discovered units
has come a long way since~\cite{jansen+etal_icassp13}.
But the evaluation above only considers two particular combinations of acoustic unit discovery and symbolic word segmentation models (rows 1 and 8).
These could be combined in different configurations.

Table~\ref{tbl:wordseg_val} shows token $F_1$ scores on Buckeye development data where different symbolic word segmentation models are applied on top of different acoustic unit discovery approaches.
The best overall $F_1$ of 24.1\% is achieved by applying DPDP AE-RNN word segmentation on DPDP CPC+$K$-means acoustic units (this is row~8 in Table~\ref{tbl:wordseg_buckeye}).
Looking only at the word segmentation models (i.e.\ comparing columns) we see that the DPDP AE-RNN performs better than the transition probability~\cite{saksida+etal_develsci17} and adaptor grammar~\cite{johnson+goldwater_naacl09} models, irrespective of which noisy acoustic unit tokenization it is applied on.

This raises the question: is it necessary to apply DPDP for acoustic unit discovery, or is it sufficient to just apply the DPDP AE-RNN directly on merged cluster indices from some VQ model?
Stated differently: is the combination of the two DPDP models important?
We see that the combination is indeed important since it gives the best overall performance (24.1\%).
But we also see that applying the DPDP AE-RNN directly on merged cluster indices from the CPC+$K$-means model---without using DPDP---comes close~(22.5\%).

Comparing the DPDP VQ-VAE and DPDP CPC+$K$-means models, we also see the importance of training the self-supervised scoring network on substantial amounts of data: the VQ-VAE is trained on roughly 15~hours of data while the CPC+$K$-means model is trained on 6k hours (\S\ref{sec:dpdp_cpc_intermediate}). The latter leads to substantially better word segmentation scores.\footnote{As a further comparison, note that the SCPC~\cite{bhati+etal_interspeech21} in Table~\ref{tbl:wordseg_buckeye} is also trained on the same 15-hour English dataset from ZeroSpeech 2019 as the VQ-VAE in Table~\ref{tbl:wordseg_val}. On the Buckeye development data, this SCPC achieves a word boundary $F_1$ and $R$-value of respectively 33.0\% and 45.6\%. In comparison, the DPDP AE-RNN on DPDP VQ-VAE achieves  23.2\% and 38.0\%.}

\subsection{Comparing to existing approaches on non-English data} 
\label{sec:experiments_nonenglish}

Although the systems above were developed and tested on different corpora, 
they were still developed and tested on the same language (English).
It is therefore unclear how the DPDP system will perform on an unseen zero-resource language when it is applied with the same hyperparameters.
For the final quantitative evaluation, I apply the chained DPDP system without alteration to French, Mandarin, German and Wolof 
data and compare it to Track 2 submissions to ZeroSpeech 2017 and 2020~\cite{dunbar+etal_asru17,dunbar+etal_interspeech20}.
Because the same data was used in the two challenges, this is one of the most comprehensive 
word segmentation benchmarks.
As scoring networks on the non-English data here, I still use the CPC-big model trained on English (\S\ref{sec:dpdp_cpc_intermediate}).

Results are shown in Table~\ref{tbl:zs2017}.
I use the ZeroSpeech evaluation suite\footnote{{\scriptsize \tt \url{https://github.com/zerospeech/zerospeech2020}}} for calculating word boundary and token scores here.
The challenge baseline system~\cite{jansen+vandurme_asru11} is an unsupervised term discovery system that optimizes for precision and therefore achieves very poor recall.
The challenge topline system is an adaptor grammar applied to phonemic transcriptions of the training data.
I only compare to a relevant subset of submitted systems.\footnote{All submission results are available at {\tt \scriptsize \url{https://zerospeech.com}}.}
With the exception of token $F_1$ on German, we
see that the DPDP system gives the best performance on the word boundary $F_1$ and word token $F_1$ scores in all other cases. 
On Mandarin, the token $F_1$ score is improved by more than 14\% absolute over the previous best result.\footnote{I should note that the DPDP system implicitly makes use of a large amount of English data for acoustic unit discovery, while the other systems only use the training set of the respective languages.
Through cross-lingual transfer, this could provide a benefit to the DPDP system, but it could also be detrimental since it never sees any within-language data for acoustic feature learning.}

\begin{table}[!t]
    \mytable
    \caption{Word segmentation results (\%) for a selection of systems from ZeroSpeech 2017 and 2020.} 
\begin{tabularx}{\linewidth}{@{}LS[table-format=2.1]S[table-format=2.1]S[table-format=2.1]S[table-format=2.1]@{}}
    \toprule
    & \multicolumn{3}{c}{Word boundary} & {Token} \\
    \cmidrule(lr){2-4} \cmidrule(l){5-5}
    Model & {Prec.} & {Rec.} & {$F_1$} & {$F_1$} \\
    \midrule
    \textit{\underline{French:}} \\
    Baseline: Sparse term discovery~\cite{jansen+vandurme_asru11} & 32.5 & 0.6 & 1.2 & 0.0 \\
    ES-KMeans~\cite{kamper+etal_asru17} & 37.0 & 52.2 & 43.3 & 6.3 \\
    Probabilistic DTW~\cite{rasanen+blandon_interspeech20} & 31.6 & \ubold 86.4 & 46.3 & 5.1 \\
    Self-expressing autoencoder~\cite{bhati+etal_interspeech20} & 34.0 & 83.9 & 48.4 & 8.3 \\
    DPDP AE-RNN on DPDP CPC+$K$-means & \ubold 49.8 &  57.9 & \ubold 53.5 & \ubold 12.2 \\
    \addlinespace
    Topline adaptor grammar on phonemes & 83.1 & 89.3 & 86.1 & 57.0 \\
    
    \addlinespace
    \textit{\underline{Mandarin:}} \\
    Baseline: Sparse term discovery~\cite{jansen+vandurme_asru11} & 54.3 & 1.3 & 2.5 & 0.2 \\
    ES-KMeans~\cite{kamper+etal_asru17} & 42.6 & 75.6 & 54.5 & 8.1\\
    Probabilistic DTW~\cite{rasanen+blandon_interspeech20} & 34.2 & 87.4 & 49.2 & 4.4 \\
    Self-expressing autoencoder~\cite{bhati+etal_interspeech20} & 36.5 & \ubold 91.9 & 52.2 & 12.1  \\
    DPDP AE-RNN on DPDP CPC+$K$-means & \ubold 66.2 & 70.7 & \ubold 68.3 & \ubold 26.3 \\
    \addlinespace		
    Topline: Adaptor grammar on phonemes & 66.2 & 100 & 79.7 & 34.9 \\
    
    \addlinespace
    \textit{\underline{{German:}}} \\
    {Baseline: Sparse term discovery~\cite{jansen+vandurme_asru11}} & 36.3 & 1.5 & 2.8 & 0.4 \\
    {ES-KMeans~\cite{kamper+etal_asru17}} & 42.9 & 66.9 & 52.3 & \ubold 14.5 \\
    {Probabilistic DTW~\cite{rasanen+blandon_interspeech20}} & 24.6 & \ubold 85.2 & 38.2 & 2.9 \\
    {Self-expressing autoencoder~\cite{bhati+etal_interspeech20}} & 27.1 & 82.8 & 40.9 & 7.5 \\
    {DPDP AE-RNN on DPDP CPC+$K$-means} & \ubold 50.5 & 61.7 & \ubold 55.6 & 9.0 \\
    \addlinespace       
    {Topline: Adaptor grammar on phonemes} & 70.0 & 98.3 & 81.8 & 50.5 \\
    
    \addlinespace
    \textit{\underline{{Wolof:}}} \\
    {Baseline: Sparse term discovery~\cite{jansen+vandurme_asru11}} & 49.9 & 1.4 & 2.7 & 0.2 \\
    {ES-KMeans~\cite{kamper+etal_asru17}} & 50.8 & 55.0 & 52.8 & 10.9 \\
    {Probabilistic DTW~\cite{rasanen+blandon_interspeech20}} & 35.2 & 48.0 & 40.6 & 4.2 \\
    {Self-expressing autoencoder~\cite{bhati+etal_interspeech20}} & 39.9 & \ubold 84.7 & 54.2 & 14.8 \\
    {DPDP AE-RNN on DPDP CPC+$K$-means} & \ubold 63.1 & 56.5 & \ubold 59.6 & \ubold 15.0 \\
    \addlinespace       
    {Topline: Adaptor grammar on phonemes} & 81.3 & 93.2 & 86.9 & 60.2 \\
    
    \bottomrule
\end{tabularx}
\label{tbl:zs2017}
\end{table}	

Despite the DPDP system's improvements, it is clear on both languages that there is still a long way to go to get to the topline performance; 
on French, German and Wolof, the best scores are still far from the 
word token $F_1$ score achieved by the idealized topline system that segments phonemic input.
In work done concurrently with this current paper (but only appearing after this paper's preprint), an extension of an earlier  nonparametric Bayesian model to speech was proposed~\cite{algayres+etal_tacl22}; despite also achieving improvements over previous ZeroSpeech sub\-missions, their approach is also still far from the topline.

\begin{figure*}[!t]
    \centering
    \includegraphics[scale=0.55]{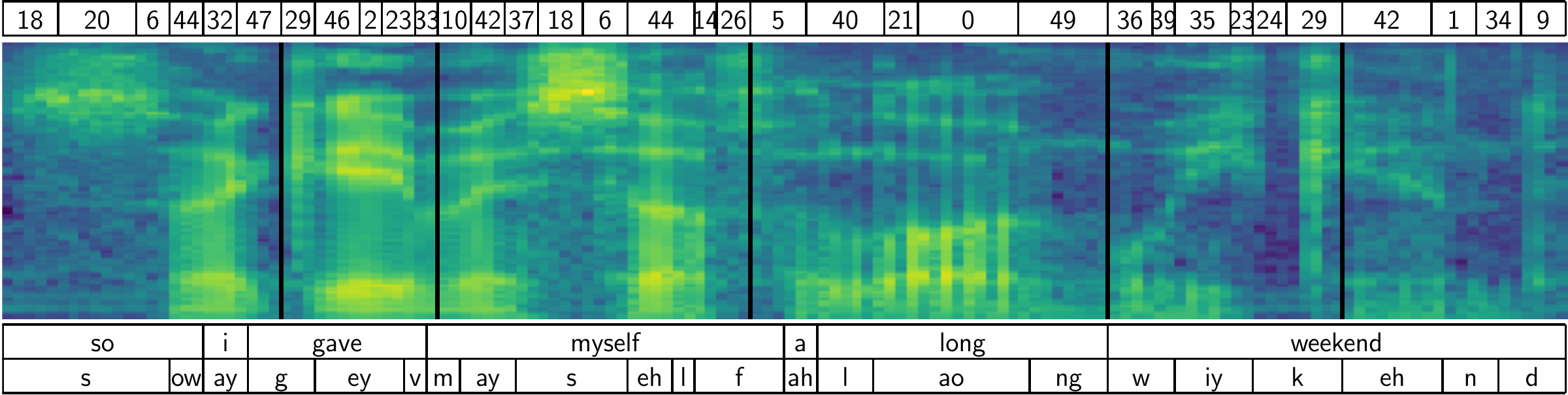}    
    \caption{An example segmentation from the chained DPDP system. The codes inferred from the DPDP CPC+$K$-means model are shown at the top. The vertical black lines on the spectrogram indicate where the DPDP AE-RNN places word boundaries. Ground truth word and phone boundaries are shown at the bottom.}
    \label{fig:dpdp_ex}
\end{figure*}

\begin{figure}[!t]
    \centering
    \includegraphics[scale=0.55]{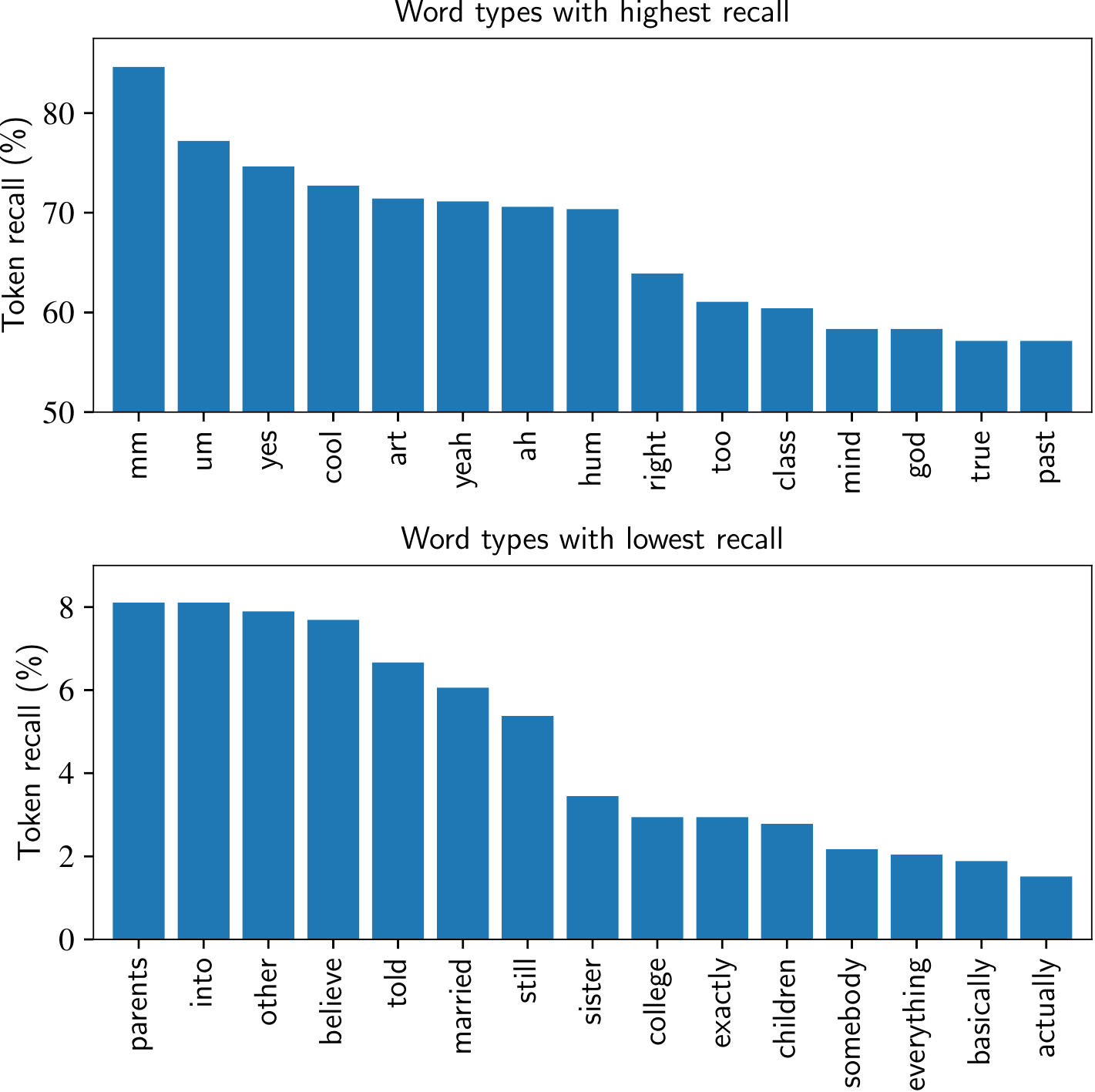}
    \caption{Word token recall from the DPDP system for individual word types.
        The best (top) and poorest recalled words (bottom) are shown.
        Note the differences in scales on $y$-axes of the two plots.}
    \label{fig:per_word_recall_best}
\end{figure}

\subsection{Qualitative analysis}
\label{sec:experiments_qualitative}

The similar performance of the DPDP system compared to state-of-the-art CPC-based models on English (\S\ref{sec:experiments_english}), and its superior performance to other systems on the non-English ZeroSpeech 
benchmarks (\S\ref{sec:experiments_nonenglish}), show that the idea of bottom-up acoustic unit discovery followed by symbolic word segmentation should not be discarded.
However, although word token $F_1$ scores have improved from around 5\% (as in~\cite{jansen+etal_icassp13}) to around 25\% (Tables~\ref{tbl:wordseg_buckeye} and~\ref{tbl:zs2017}), absolute scores are still low.
What are the characteristics of the words that are correctly segmented? And which words do the DPDP system still struggle to segment?

Figure~\ref{fig:per_word_recall_best} shows the token recall for the 15 word types that are most often segmented correctly (top) and the words that are incorrectly segmented most often (bottom).\footnote{To get the recall for a particular word type, the number of correctly segmented word tokens (where both boundaries are correct without an intermediate prediction) is divided by the total number of tokens of this~type.}
Several of the correctly segmented words are shorter filler words: ``mm'' and ``um'' have the highest recall.
In contrast, the word types with the lowest recall are longer words.
Several of these end on \mbox{[l iy]}, and a manual review of segmentations reveals that a boundary is often placed just before these phones.
In the segmentation example in Figure~\ref{fig:dpdp_ex} we see a similar type of erroneous segmentation where ``weekend'' is segmented into two segments: [w iy k] and [eh n d].

Another indication that the DPDP system has a bias towards shorter words is if you look at the duration of correctly segmented word tokens: 231~ms with a standard deviation of 115~ms.
Compare this to the same statistics for the ES-KMeans: 286~ms $\pm$ 124~ms.
There is clearly large variation in these durations, but it is evident that the DPDP system's segmentation leans towards shorter words.

%

\section{Conclusion, Discussion and Future Work}

I have described a model for unsupervised phone segmentation and a model for unsupervised symbolic word segmentation.
I framed both as instances of a duration-penalized dynamic programming (DPDP) procedure, where a self-supervised neural scoring function is combined with a penalty term that encourages longer segments.
I chained the two models to do word segmentation from speech and compared this to existing direct whole-word and joint self-supervised approaches on standard benchmarks.
The results showed that purely bottom-up phone discovery followed by symbolic segmentation---the methodology exemplified in the chained DPDP system---performs competitively to state-of-the-art models.
This includes a comparison to
two recent joint models that extend contrastive predictive coding (CPC) for segmentation~\cite{bhati+etal_interspeech21,cuervo+etal_arxiv21}.
My argument is not that the chained methodology would in general be
superior to joint modelling, but rather that it still has a place for further investigation for the task of speech segmentation. (It is even possible that the DPDP approach could be extended and improved through joint training in the future.)

Despite the encouraging results, absolute word segmentation scores remain low: on French, Mandarin, German and Wolof
benchmarks, the best approaches are still far from a topline system where symbolic word segmentation is performed on transcribed phoneme sequences.
On English, it is also interesting how similar the scores from the DPDP system are 
compared to the joint CPC-based approaches---despite the big differences in underlying methodology.
Are we reaching the limits of what can be learned purely from unlabelled acoustic data?
If so, how can we bridge the gap that remains between these systems and the idealized topline systems?

I believe that other sources of top-down information are needed.
The qualitative analyses in this paper revealed a bias towards segmenting shorter (often filler) words.
One approach would be to incorporate top-down information from a sparse term discovery system tailored towards discovering recurring but longer words~\cite{park+glass_taslp08,jansen+vandurme_asru11,rasanen+blandon_interspeech20}, which could be used to bootstrap segmentation.
Another approach would be to incorporate information from another modality; we know that infants have access to cross-situational cues from different modalities that can aid word learning~\cite{rasanen+rasilo_psych15}.
In analogy, top-down information from visually grounded speech (VGS) models~\cite{harwath+etal_nips16,gelderloos+chrupala_coling16,scharenborg+etal_icassp18} could be used to help with segmenting
particular words.
In the analysis of~\cite{olaleye+etal_arxiv22},
words that are accurately localized with an unsupervised VGS system include several words that are longer than those in the error analysis here (\S\ref{sec:experiments_qualitative}).
So the visual grounding information could provide a complementary segmentation signal (see~\cite{peng+harwath_interspeech22} for very recent work on this).
However, incorporating top-down signals for segmentation comes with its own challenges:~\cite{sanabria+etal_insights21} shows that speech segmentation is difficult
even in a setting where text transcriptions or translations
are used as an additional modality.

Apart from these future directions, there are two particular shortcomings of the DPDP system that need to be addressed.
First, both DPDP approaches use a very coarse duration model, where the same duration penalty is applied irrespective of the acoustic unit or word segment under consideration.
To address this,
probabilistic models that explicitly model per-unit durations could be integrated or adapted to the DPDP approaches proposed here~\cite{murphy_notes02,johnson+willsky_jmlr13}.
Secondly, in contrast to some older speech segmentation models~\cite{lee+etal_taslp15,kamper+etal_csl17}, the DPDP autoencoding recurrent neural network (AE-RNN) does not infer an explicit lexicon---it can predict word boundaries, but word categories are not learned (categories are also not learned in~\cite{bhati+etal_interspeech21,cuervo+etal_arxiv21}).
I did initial experiments where the latent DPDP AE-RNN representations are clustered with $K$-means, and also tried to then re-segment the acoustic units while taking this clustering into account, but this hurt overall segmentation performance.
Building up a lexicon within the DPDP AE-RNN will therefore require further investigation.

\section*{Acknowledgements}

Thanks go to Robin Algayres for help with the ZeroSpeech tools; Benjamin van Niekerk, Leanne Nortje, Larissa Tredoux, Urs de Swart and Matthew Baas for helpful suggestions; and Roger Moore for pointing out the links to the much earlier work done at BBN.
This work is supported in part by the National Research Foundation of South Africa (grant no. 120409).

\bibliography{mybib}

\end{document}